\documentclass[runningheads]{llncs}

\usepackage{cite}
\usepackage{amsfonts, amssymb, amsmath}

\usepackage{booktabs}

\usepackage[T1]{fontenc}
\usepackage{graphicx}
\makeatletter
\renewcommand{\section}{\@startsection{section}{1}{\z@}%
  {1.5ex plus .2ex minus .1ex}%   space *before* the title
  {0.8ex plus .1ex}%             space *after* the title
  {\normalfont\large\bfseries}}

\renewcommand{\subsection}{\@startsection{subsection}{2}{\z@}%
  {1.0ex plus .15ex minus .08ex}% before
  {0.6ex plus .08ex}%            after
  {\normalfont\normalsize\bfseries}}

\renewcommand{\subsubsection}{\@startsection{subsubsection}{3}{\z@}%
  {0.8ex plus .1ex minus .05ex}% before
  {0.4ex plus .05ex}%            after
  {\normalfont\normalsize\itshape}}
\makeatother

\usepackage{enumitem}
\setlist{nosep,leftmargin=*,topsep=2pt,parsep=0pt,itemsep=0pt}

\begin{document}
\title{Multi-task parallelism for robust pre-training of graph foundation models on multi-source, multi-fidelity atomistic modeling data}
\author{Massimiliano Lupo Pasini\inst{1}\orcidID{0000-0002-4980-6924} \and
Jong Youl Choi\inst{1}\orcidID{0000-0002-6459-6152} \and
Pei Zhang\inst{1}\orcidID{0000-0002-8351-0529} \and Kshitij Mehta \inst{1} \orcidID{0000-0002-9714-9981} \and Rylie Weaver \inst{1} \orcidID{0000-0002-5205-7602} \and Ashwin M. Aji \inst{2} \and Karl W. Schulz \inst{2} \and Jorda Polo \inst{2} \and Prasanna Balaprakash \inst{1}}
\authorrunning{M. Lupo Pasini et al.}
\institute{Oak Ridge National Laboratory, Oak Ridge, TN, 37831, USA 
\email{\{lupopasinim,choij,zhangp1,mehtakv,weaverre,pbalapra\}@ornl.gov}\\
\and
Advanced Micro Devices Research, USA\\
\email{\{ashwin.aji,karl.schulz,jorda.polo\}@amd.com}}
\maketitle              %
\begin{abstract}
Graph foundation models using graph neural networks promise sustainable, efficient atomistic modeling. To tackle challenges of processing multi-source, multi-fidelity data during pre-training, recent studies employ multi-task learning, in which shared message passing layers initially process input atomistic structures regardless of source, then route them to multiple decoding heads that predict data-specific outputs. This approach stabilizes pre-training and enhances a model's transferability to unexplored chemical regions. Preliminary results on approximately four million structures are encouraging, yet questions remain about generalizability to larger, more diverse datasets and scalability on supercomputers. We propose a multi-task parallelism method that distributes each head across computing resources with GPU acceleration. Implemented in the open-source HydraGNN architecture, our method was trained on over 24 million structures from five datasets and tested on the Perlmutter, Aurora, and Frontier supercomputers, demonstrating efficient scaling on all three highly heterogeneous super-computing architectures.

\end{abstract}

\section{Introduction}

As the use of artificial intelligence (AI) and machine learning (ML) is expanding and disseminating for atomistic modeling applications, valid concerns have been recently raised with respect to the amount of task-specific data and computational resources needed to develop AI/ML models that achieve desired accuracy. In response, 
a new paradigm has been adopted to develop graph foundation models (GFMs) for atomistic modeling applications in two steps. The first step, called pre-training, consists in training a graph neural network (GNN) model on large volumes of generic, task-agnostic atomistic data using supercomputing facilities. The second step, called fine-tuning, customizes the pre-trained GNN on specific applications of interest called downstream tasks. While the pre-training phase is typically data- and compute-intensive, it is performed only once and results in a drastic reduction of data volume and computational resources needed for task-specific fine-tuning. % to a diverse set of problems. 

Although several efforts have already developed GFMs for atomistic modeling applications \cite{takeda2023multimodal, 10.1145/3624062.3626081, beaini2024towards, sypetkowski2024on, barrosoluque2024openmaterials2024omat24} by training GNNs on first-principles data, these works used a single dataset of choice for pre-training. Consequently, the pre-trained models are limited to a relatively narrow set of compounds, e.g., either organic or inorganic, and fail to provide robust GFMs that can support a broad and diverse set of atomistic scale applications. 

To expand model generalizability, pre-training must process large volumes of atomistic data on diverse sets of atomistic structures that originate from different sources and differ in the chemical composition and atomistic configuration. Achieving such a level of generality for a pre-trained GFM requires aggregating datasets that differ in(i) approximation theories adopted to run first-principles calculations such as density functional theory (DFT) and coupled clusters singles and doubles (CCSD), and (ii) parameterizations for a given approximation theory (e.g., the exchange correlation functional in DFT). 

The heterogeneity across multi-source, multi-fidelity data poses significant challenges for GFM pre-training, which is prone to numerical instabilities as illustrated in \cite{LupoPasini2025HydraGNN}. To address this challenge, recent works have explored the use of multi-task learning (MTL) to jointly pre-train GNNs using multi-source datasets labeled with different DFT settings \cite{Jacobson2023, shoghi2024from}. In this approach, the shared layers of the MTL architecture learn relevant features during pre-training that are common to all datasets and boost transferability across diverse classes of compounds.  Each MTL output head focuses on features specific to its assigned dataset, which facilitates the fine-tuning of the GFM across a broad range of downstream tasks. 
Early results have illustrated the efficacy of MTL in improving both pre-training stability and model transferability for model usage in regions of the chemical space that were not covered during the pre-training. 
Inspired by these promising results, the authors of \cite{shiota2024tamingmultidomainfidelitydata} expanded the approach to a broader set of datasets and compounds for GFM development, re-confirming the success of MTL in stabilizing the pre-training of GFMs on multi-source, multi-fidelity data.  
However, the studies mentioned above were limited to datasets of moderate volume (at most 3 million atomistic structures), and overlooked the scalability challenges, which are crucial given the increased amount of open-source datasets continuously released.

The two primary contributions of our work are aggregating large volumes of available multi-source, multi-fidelity data at an unprecedented degree of heterogeneity, and the development of a new model parallelism method called multi-task parallelism, which distributes different output decoding heads of an MTL architecture across distributed computing resources. Our multi-task parallelism approach distributes MTL heads across multiple GPUs and allows for all heads to concurrently process the data propagated forward by the shared layers of the GNN architecture. 
Allowing MTL heads to be distributed across compute resources addresses the scalability limitations of traditional MTL when a large model does not fit in GPU memory. This allows for scaling up the pre-training of GFMs as the number of datasets increases by simply increasing the number of GPUs used to host the new MTL heads for the new datasets. 
In addition, we combine multi-task parallelism with distributed data parallelism (DDP) to form a 2D parallelization that enables efficient processing of large volumes of multi-source, multi-fidelity data. 
We illustrate the performance of our approach by pre-training GFMs on 5 datasets that amount to over 24 million atomistic structures. The numerical results confirm what was already noticed by previous works at a smaller scale, namely that MTL effectively balances the GFM pre-training on multi-source, multi-fidelity data. Strong and weak scaling tests performed on the Perlmutter petascale system, the Aurora exascale system, and the Frontier exascale system assessed the scalability of our method using up to 640 GPUs on Frontier and Perlmutter and 1920 GPUs on Aurora.

The remainder of the paper is organized as follows. In Section \ref{relatedwork} we discuss existing work on addressing the challenge of inconsistent multi-source, multi-fidelity datasets in atomistic scale applications to train deep learning models. Section ~\ref{sec:background} provides background on the HydraGNN graph neural network architecture. We discuss our MTL approach in Section ~\ref{sec:OurContribution} along with convergence and scaling results in Section ~\ref{sec:results}. We provide a conclusion and future work in Section ~\ref{sec:conclusion}.

\section{Related Work}
\label{relatedwork}

We survey two lines of prior work: (i) learning from \textit{multi-source, multi-fidelity} data and (ii) \textit{scalable GNN training}.

\subsection{Training on multi-source, multi-fidelity data}
No single dataset or approximation theory offers equally high accuracy across both organic and inorganic compounds. Thus, developing generalizable GFMs requires training on multi-source, multi-fidelity data.
Our datasets do not describe the same physical systems at different fidelity levels but instead span different atomistic domains. Therefore, conventional multi-fidelity learning methods such as Correlation Alignment for domain adaptation (CORAL) \cite{sun2016deep} and Maximum Mean Discrepancy (MMD) \cite{gretton2012kernel} cannot be applied.

Early transfer-learning studies pre-train on low-level DFT and fine-tune on high-level CCSD for small organic molecules \cite{Smith2019,Smith2020ANI1ccx}; the approach was later expanded to several datasets but still a narrow chemical space \cite{allen2024foundation}.  Shiota et al.\ proposed “total-energy alignment’’ (ICEA + AEC) to merge inconsistent organic and inorganic data before single-task training \cite{shiota2024tamingmultidomainfidelitydata}, yet the method needs reference atomic energies and may not generalise to many datasets.  Zhang et al.\ instead used \textit{multi-task learning} (a shared encoder with per-dataset heads) and achieved better accuracy and transferability \cite{zhang2024dpa2largeatomicmodel}.  Their study, however, covered only $\sim$4 M structures—half self-generated—and ignored HPC concerns, leaving open questions about scalability.

\subsection{Distributed GNN training}
Most large-graph frameworks combine data partitioning with distributed data parallelism (DDP): AliGraph \cite{aligraph}, FlexGraph \cite{flexgraph}, DistDGL \cite{distgnn} and Hybrid-DistGNN \cite{hybrid_dist_gnn}.  Intra-layer model-parallel schemes such as NeuGraph and GNNAdvisor tackle sparse kernels, akin to tensor parallelism in LLMs. However, these tools are suited to train GNNs on monolithic graphs with millions of nodes, whereas atomistic workloads involve millions of \textit{small} graphs, each with a few hundreds of nodes.  

\section{Background}
\label{sec:background}

We perform our study using the HydraGNN open-source graph convolutional neural network architecture~\cite{hydragnn3, usermanual_hydragnn, LupoPasini2025HydraGNN}. Some of the important features of HydraGNN for scalable training are as follows.
\begin{itemize}
\item MTL capabilities to process multi-source, multi-fidelity data;
\item Object-oriented programming capabilities to use different MPNN layers, which allows flexible switching between different message policies based on the scientific needs of the specific application at hand, as well treating the MPNN layer as a tunable categorical hyperparameter with hyper-parameter optimization;
\item Invariant and equivariant features that reduce computational redundancy and time-to-solution, therefore contributing to energy saving;
\item Scalable input/output (I/O) data management techniques to efficiently scale the training of GNN models on millions of data samples using thousands of GPUs on supercomputing facilities; and
\item Portable capabilities that allow conveniently running the GNN training on diverse computing platforms with different hardware and software specifications.
\end{itemize}

HydraGNN supports several MPNN layers, including some that build invariant and/or equivariant features. 
Invariant and equivariant message passing layers are a specialized type of MPNN layers designed to account for certain symmetries in the input data. It ensures that the output of the layer transforms in a predictable and consistent way under transformations of the input. This is particularly useful in atomistic materials modeling because symmetries play a crucial role in determining properties such as energy per atom and atomic forces.

HydraGNN efficiently stores and reads large training datasets using the ADIOS~\cite{adiosSoftwareX} scientific data management library by serializing and storing data samples into its custom scientific data format, and reading them in parallel during the training process. ADIOS provides an efficient storage solution and helps obtain high I/O bandwidth when large volumes of data are ingested into HydraGNN.
HydraGNN uses a two-pronged approach in which all data samples are read from their ADIOS files into DDStore~\cite{ddstoreChoi}, an in-memory data store that manages a distributed cache across all MPI processes. DDStore provides low-latency data access operations to transfer data between processes. When an epoch completes, a process requests the next batch of data from DDStore, which transparently obtains it from the memory of a remote process using one-sided operations. It completely circumvents accessing the file system and provides a high-throughput communication mechanism for obtaining a data batch.

\section{Our Contribution}
\label{sec:OurContribution}

The novelty of our approach consists of two main contributions. 
First, we aggregated large volumes of available multi-source, multi-fidelity data at an unprecedented degree of heterogeneity and diversity across 5 distinct datasets. The number of atomistic structures used as independent data samples after aggregating the 5 datasets amounts to over 24 million, almost 6x larger than what was used in \cite{zhang2024dpa2largeatomicmodel}. We consistently aligned the energy per atom values across all the datasets, and used the aligned data to assess the stabilization provided by MTL during the GFM pre-training.

Second, we developed and implemented a new model parallelism method called multi-task parallelism specifically for MTL, which involves distributing the MTL output heads assigned to different datasets across different GPUs. This allows for concurrent executions of the forward and backward passes between MTL heads that work on distinct portions of the data. More details about the implementation of our multi-task parallelism approach are provided in Section \ref{subsec:mtl}.

\subsection{Data preparation}

Using large datasets for GFM pre-training is expected to enhance generalizability and ensure resilience to data variance issues that typically arise during downstream tasks.
To this end, we aggregated five open-source atomistic materials modeling datasets that are extremely diverse in terms of chemical composition, atomistic configurations, and number of atoms in the system. 
These datasets, are: \texttt{ANI1x}, \texttt{QM7-X}, \texttt{Transition1x}, \texttt{MPTrj}, and \texttt{Alexandria}.
\begin{itemize}
\item \texttt{ANI1x} dataset \cite{Smith2020ANI1ccx} consists of 4,956,005 density functional theory (DFT) calculations of energies and forces for atomistic structures derived from up to 57 thousand distinct molecular configurations containing the C, H, N, and O natural elements. 
\item \texttt{QM7-X}  \cite{qm7x} is a comprehensive dataset of 42 physicochemical properties for approximately 4.2 million equilibrium and non-equilibrium structures of small organic molecules with up to seven non-hydrogen atoms from the C, N, O, S, Cl chemical elements.
\item \texttt{Transition1x} dataset \cite{schreiner2022transition1x} contains 9.6 million DFT calculations of energies and forces of molecular configurations on and around reaction pathways at the $\omega$B97x\/6–31G(d) level of theory and contains the C, H, N, O, F, S, Cl, P, Br, I, Li, Na, and K natural elements. 
\item \texttt{MPTrj} \cite{mptrj}: the version of the dataset from 2020 provides over 1.5 million single point DFT calculations of energies and forces for near-equilibrium atomistic structures of inorganic materials, and covers over 60 natural elements. 
\item \texttt{Alexandria} \cite{alexandriapbepaper, alexandriapaper, alexandria2dpaper, alexandria3dpaper, alxandriaphononpaper} dataset provides DFT calculations of energies and forces for over 4.9 million equilibrium and non-equilibrium atomistic structures of inorganic materials.
\end{itemize}

Each dataset is unique for the chemical compositions and the number of atoms in the atomistic structures of the compounds described. 
In total, the data used for training, validating, and testing our GFM consisted of over 24 million atomistic structures that cover over two-thirds of the natural elements of the periodic table. 
These datasets were pre-processed using the ADIOS scientific data management library into a common format for efficient storage and I/O, as discussed in Section \ref{sec:background}.

\begin{figure}[ht]
   \centering
   \includegraphics[width=0.8\columnwidth]{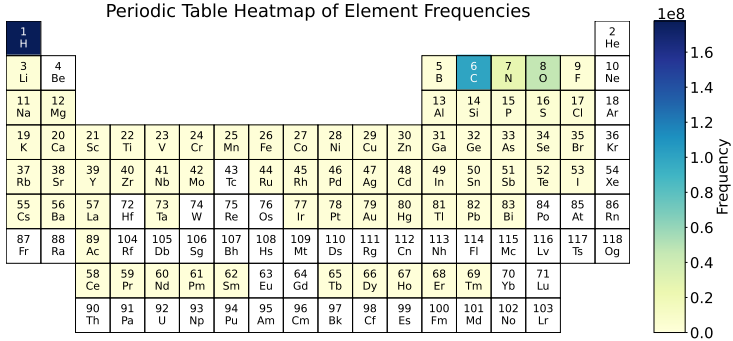}
   \caption{Heatmap that describes the frequency of occurrence of each element of the periodic table across data sampled resulting from the aggregation of the datasets \texttt{ANI1x}, \texttt{QM7-X}, \texttt{MPTrj}, \texttt{Alexandria}, and \texttt{Transition1x}.}
\label{fig:heatmap}
\end{figure}

Figure \ref{fig:heatmap} provides the heatmap that illustrates the frequency of occurrence of each element of the periodic table across the entire dataset that results from the aggregation of the datasets \texttt{ANI1x}, \texttt{QM7-X}, \texttt{Transition1x}, \texttt{MPTrj}, and \texttt{Alexandria}.

\subsection{Two-level hierarchical MTL architecture}

\begin{figure}
    \centering
    \includegraphics[width=0.7\columnwidth]{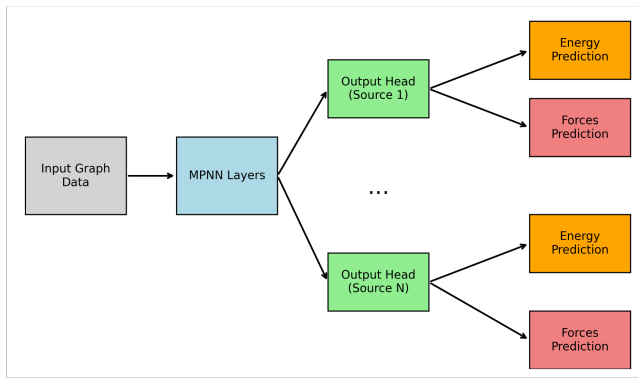}
    \caption{HydraGNN architecture that implements a two-level MTL approach. The first level of MTL splits the architecture into multiple output decoders to process data coming from different sources, whereas the second level of MTL further splits each output decoder into two sub-decoders for the simultaneous prediction of energies and atomic forces.}
\label{fig:mtlgcnn}
\end{figure}

In our approach, we need to apply MTL both to process multi-source multi-fidelity data as well as to simultaneously predict multiple target properties, namely energy per atom and atomic forces. 
To this aim, we apply a two-level hierarchical MTL approach. At the first MTL level, the HydraGNN architecture splits into multiple branches, each dedicated to processing data from a specific dataset. At the second level, MTL is applied again to split each branch into two output heads, one dedicated to predicting the energy per atom and the other to predicting the atomic forces. We illustrate the resulting HydraGNN architecture with the two-level hierarchical MTL in Figure \ref{fig:mtlgcnn}. 
As the number of datasets increases, this approach can be scaled up by adding more data-specific MTL branches to the setup, but is limited by the memory capacity of GPU compute units as the model size increases.

\subsection{Multi-task parallelism}
\label{subsec:mtl}

\begin{figure}
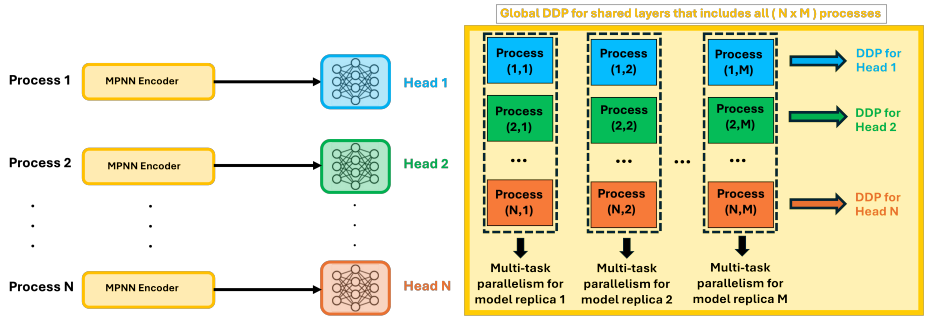

\includegraphics[width=0.5\columnwidth]{figures/MTL\_DDP1.png}
\includegraphics[width=0.5\columnwidth]{figures/MTL\_DDP2.png}
\caption{{\it Left:} Partitioning of one single HydraGNN model replica across N processes. Each process is mapped to one NVIDIA A100 GPU on NERSC-Perlmutter, one AMD Mi250X GCD on OLCF-Frontier, and one Intel Data Center GPU Tile on ALCF-Aurora. {\it Right:} Organization of processes into sub-process groups using \texttt{torch.DeviceMesh}. One global group is used to perform DDP to synchronize the gradients associated with the MPNN encoder parameters. $N$ sub-process groups (each one with $M$ processes) perform local DDPs to synchronize the gradients associated with parameters of the MTL output decoding heads. }
\label{fig:torchmesh}
\end{figure}

Our GFM model design incorporates dataset-specific branches which naturally enable parallelism.
Since different MTL output decoding heads of an HydraGNN architecture process data from different sources, the forward propagation of an MTL output head does not require coordination with the other output heads. Therefore, forward propagations for each MTL head can be performed concurrently using independent processes. This motivates our proposed multi-task level parallelism, which is a specific type of model parallelism suited for MTL architectures. The idea behind multi-task parallelism is to split one multi-head HydraGNN model replica across multiple processes, as illustrated in Figure \ref{fig:torchmesh}. Each process owns a local copy of the parameters of the shared MPNN layers and a local copy of the parameters corresponding to one single MTL output decoding head. 
During the backward propagation needed to update the model parameters, each process starts independently and concurrently updating the parameters or their own respective MTL output decoding heads. Then the backpropagation continues updating the parameters of the shared MPNN layers, for which the processes need to synchronize with a collective operation to compute an average of the gradient updates.

Our multi-task parallelism becomes advantageous when the number of MTL heads—corresponding to datasets—grows such that the total parameter count exceeds GPU memory capacity. The goal of our parallelization is not to replace pipeline/tensor parallelism but to complement them. Traditional model parallelism is suited for deep or wide networks. These approaches do not address the challenge posed by a growing number of MTL heads, which is unique to multi-source, multi-fidelity data. Our method targets this gap and can be combined with other strategies if needed. 

We denote the parameter count of shared layers with $P_s$, the parameter count of heads with $P_h$, and the number of heads with $N_h$. Without parallelism, memory per GPU scales as $P_s+(P_h * N_h)$; with it, memory per GPU is reduced to $(P_s+P_h)$. This also enables larger batch sizes, reducing the need for more compute nodes under distributed data parallel (DDP) training.
We identified three regimes were different types of parallelization should be prioritized:
\begin{itemize}
\item Case 1: $ P_s >> (N_h * P_h) \rightarrow$  pipeline/tensor parallelism preferred.
\item Case 2: $ P_s << (N_h * P_h) \rightarrow$  multi-task parallelism optimal.
\item Case 3: $ Ps \sim (N_h * P_h) \rightarrow$  hybrid schemes recommended.
\end{itemize}
GNNs, especially message passing neural networks (MPNNs), tend to fall under Case 2, motivating our approach.

\subsection{Integration of multi-task parallelism with DDP}
In order to process large volumes of data during the GFM pre-training, we integrate our multi-task parallelism with distributed data parallelism by dividing processes into sub-groups and distributing the workload among them. Each sub-group is dedicated to a specific dataset and processes within each subgroup share the MTL-specific layers, while the MPNN layers are shared globally across all subgroups.
We use Pytorch's \texttt{torch.DeviceMesh}\footnote{\url{https://pytorch.org/tutorials/recipes/distributed\_device\_mesh.html}} package which allows  setting up communicators to create  process sub-groups. 
Processes within a sub-group handle the same MTL output head across different HydraGNN model replicas. We illustrate the structuring of the processes in sub-groups in Figure \ref{fig:torchmesh}.

\section{Numerical Results}
\label{sec:results}

We evaluate our approach on the three DOE supercomputing facilities dedicated to open-science research: the petascale system NERSC-Pelrmutter\footnote{\url{https://docs.nersc.gov/systems/perlmutter/architecture/}}, the exascale system OLCF-Frontier\footnote{\url{https://docs.olcf.ornl.gov/systems/frontier\_user\_guide.html}}, and the exascale system ALCF-Aurora\footnote{\url{https://www.alcf.anl.gov/aurora}}. 

For all experiments we used the best-performing HydraGNN variant found during our earlier scalable hyper-parameter search \cite{LupoPasini2025HydraGNN}. The backbone is a 4-layer Equivariant GNN (EGNN) with 866 hidden units per message-passing layer; each dataset-specific head consists of three fully-connected layers of 889 units. This single configuration—selected from fifteen candidates—serves throughout the paper for both energy- and force-prediction tasks.

\subsection{Training convergence}
The first set of numerical results compares the predictive performance of (1) training a HydraGNN model on a single dataset at a time, (2) pre-train a HydraGNN model on all datasets mixed together using one-level MTL to simultaneously predict energy per atom and atomic forces as in \cite{LupoPasini2025HydraGNN}, and (3) use a two-level MTL approach to process multi-source, multi-fidelity data with multiple branches, and each branch splits into two MTL output heads to predict energy per atom and forces for a single data-source. This amounts to a total of seven HydraGNN models trained for this set of numerical results. 
The batched stochastic optimizer used is AdamW with a learning rate set to 0.001 and a local batch size set to 128 data samples. 
The training was distributed across 128 nodes of OLCF-Frontier with a total of 1,024 DDP processes. Early stopping was applied to avoid redundant computations by further training models that would not improve their accuracy.

Tables~\ref{tab:energy-mae} and \ref{tab:force-mae} show the MAE of the energy per atom and force predictions, respectively. For each of the five datasets used in this study, we highlight the MAE of the two best performing models. While the models trained on each individual dataset perform well for in-distribution prediction accuracy on the testing portion of the same dataset they were trained on, they perform poorly on other datasets because these datasets contain atomistic structures with chemical composition and configuration that significantly differ from the ones used for the training.
Among the five datasets, \texttt{MPTrj} and \texttt{Alexandria} are particularly challenging and cause more out-of-distribution errors than the other datasets due to the fact that \texttt{MPTrj} and \texttt{Alexandria} are the only datasets that include inorganic compounds. Moreover, the chemical space covered by them is much broader than the chemical space covered by the other datasets, which results in an increased sparsity of data samples. 
The baseline GFM pre-trained with all datasets mixed together and processed by a single output decoding head, denominated as GFM-Baseline-All in the result tables, shows overall improved transferability compared to the the dataset-specific models, but with deteriorated prediction accuracy due to its limited capability to handle highly heterogeneous datasets.
In contrast, the MTL-GFM pre-training with individual output decoding heads for each dataset, denominated as GFM-MTL-ALL in the result tables, delivers outstanding performance with a combined high accuracy and robust transferability across all datasets. These results confirm that MTL is an efficient training scheme that significantly improves the accuracy and transferability of GFMs for atomistic modeling when they are trained on heterogeneous, multi-source, multi-fidelity data.

\begin{table}[htbp]
  \centering
  \caption{MAE error in energy per atom predictions of test set from the seven models on five datasets}\label{tab:energy-mae}
  {
  \begin{tabular}{|l|c|c|c|c|c|}
    \toprule
    \hline
     & ANI1x & QM7-X & MPTrj & Alexandria & Transition1x \\
    \midrule
    \hline
    Model-ANI1x & $\boldsymbol{0.0005}$   & 0.1252   & 10.7739  & 22.7022  & 0.1056 \\
    Model-QM7-X & 0.1480   & $\boldsymbol{0.0081}$  & 319.5903 & 962.2206 & 0.1230 \\
    Model-MPTrj & 0.3323   & 0.3902   & $\boldsymbol{0.0651}$    & 0.3766   & 0.3690 \\
    Model-Alexandria & 0.0721   & 0.1189   & 0.4729   & $\boldsymbol{0.0057}$   & 0.1249 \\
    Model-Transition1x & 0.1721  & 0.1957   & 19.4042  & 24.6109  & $\boldsymbol{0.0138}$ \\
    GFM-Baseline-All        & 0.0119   & 0.0655   & 0.4248   & 0.0634   & 0.1030 \\
    GFM-MTL-All  & $\boldsymbol{0.0007}$ & $\boldsymbol{0.0096}$ & $\boldsymbol{0.0627}$ & $\boldsymbol{0.0179}$ & $\boldsymbol{0.0115}$  \\
    \bottomrule
    \hline
  \end{tabular}}
\end{table}

\begin{table}[htbp]
  \centering
  \caption{MAE error in force predictions of test set from the seven models on five datasets}\label{tab:force-mae}
 {
  \begin{tabular}{|l|c|c|c|c|c|}
    \toprule
    \hline
     & ANI1x & QM7-X & MPTrj & Alexandria & Transition1x \\
    \midrule
    \hline
    Model-ANI1x & $\boldsymbol{0.0034}$   & 1.1323   & 67.9839   & 234.0164  & 0.1400   \\
    Model-QM7-X & 1.2797   & $\boldsymbol{0.0582}$   & 1558.8623 & 7286.7197 & 0.3192   \\
    Model-MPTrj & 0.0508   & 1.1734   & $\boldsymbol{0.1597}$    & $\boldsymbol{0.0039}$   & 0.1425   \\
    Model-Alexandria & 0.0508   & 1.1734   & $\boldsymbol{0.1597}$    & $\boldsymbol{0.0038}$   & 0.1425   \\
    Model-Transition1x & 0.7686   & 1.1668   & 98.0733   & 160.0776  & $\boldsymbol{0.0433}$   \\
    GFM-Baseline-All        & 0.0508   & 1.1734   & $\boldsymbol{0.1597}$   & $\boldsymbol{0.0039}$   & 0.1425   \\
    GFM-MTL-All     & $\boldsymbol{0.0074}$   & $\boldsymbol{0.0925}$   & $\boldsymbol{0.1238}$   & $\boldsymbol{0.0039}$   & $\boldsymbol{0.0388}$   \\
    \bottomrule
    \hline
  \end{tabular}}
\end{table}

\subsection{Scaling results}

\begin{figure}
    \centering
    \includegraphics[width=0.3\columnwidth]{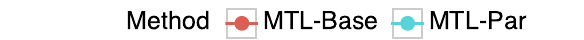}\\
    \includegraphics[width=0.3\columnwidth]{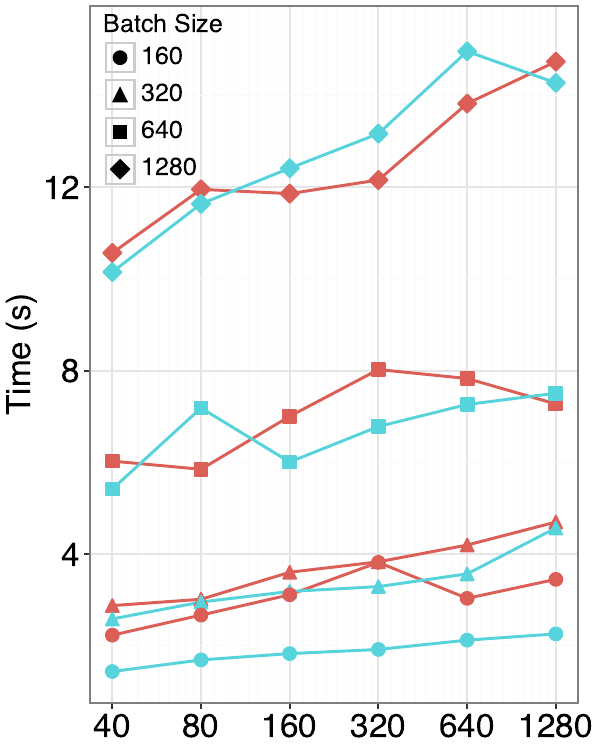}
    \includegraphics[width=0.3\columnwidth]{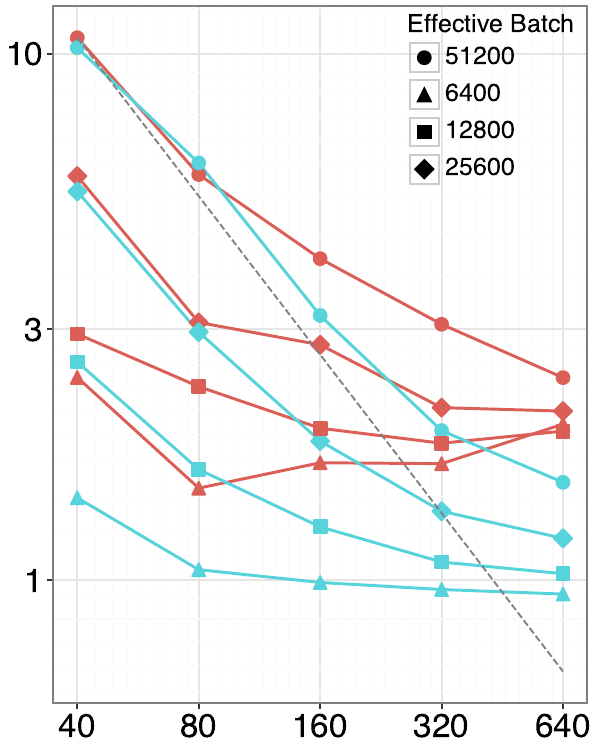}\\
    \includegraphics[width=0.3\columnwidth]{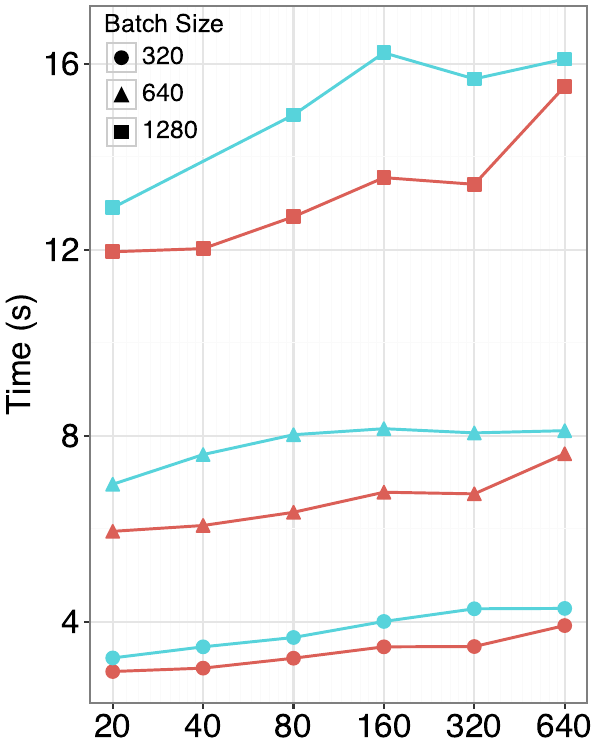}
    \includegraphics[width=0.3\columnwidth]{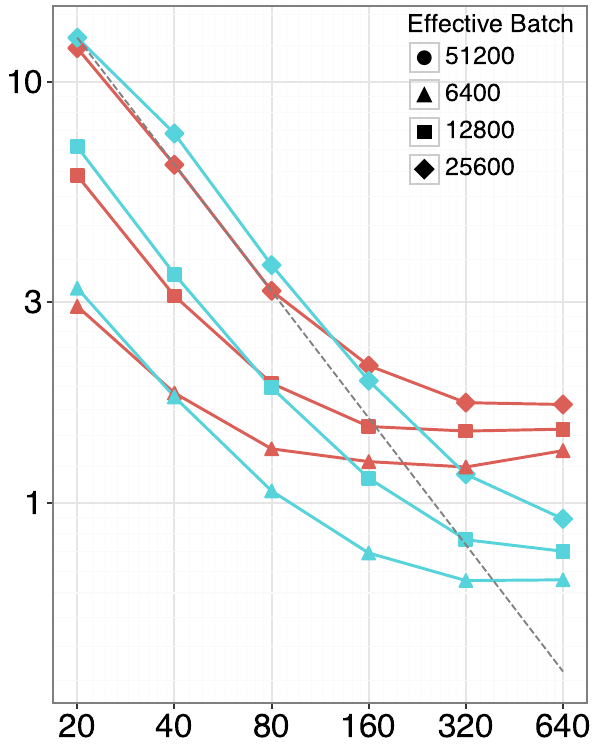}\\
    \includegraphics[width=0.31\columnwidth]{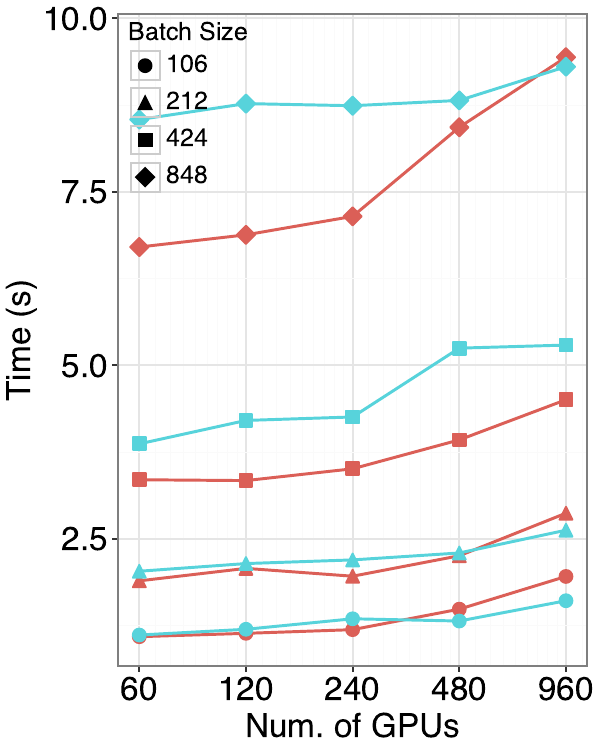}
    \includegraphics[width=0.3\columnwidth]{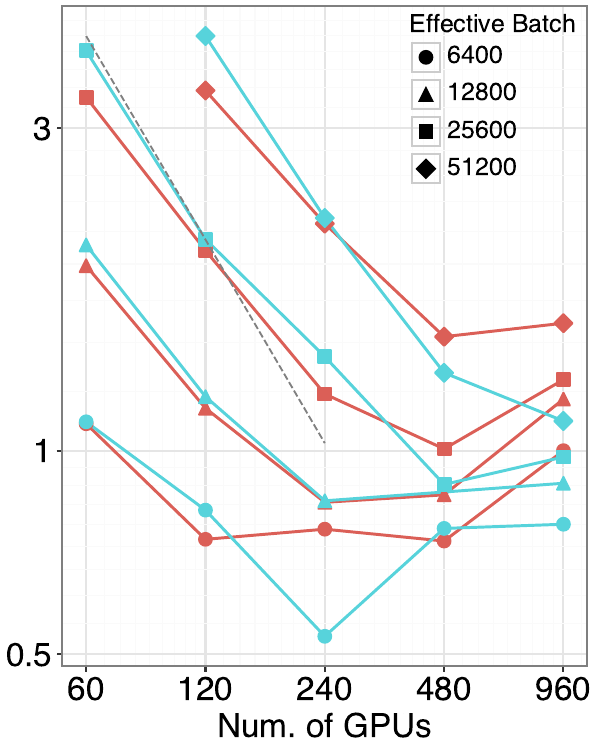}\\
    \caption{Scaling experiment results. The plots show weak scaling (left) and strong scaling (right) on Frontier (top), Perlmutter (middle), and Aurora (bottom). The y-axis shows the average total training time per epoch, including data loading, forward, and backward passes, while the x-axis indicates the number of GPUs used. For the strong scaling plots, a black dashed line is shown to indicate the ideal linear scaling trend.}
    \label{fig:scaling}
\end{figure}

Figure ~\ref{fig:scaling} shows the scaling performance of our MTL methods on all the three supercomputing systems for different batch sizes. We refer to MTL with DDP as `MTL-base', whereas MTL with DDP and task parallelism is denoted as `MTL-par'. In MTL-par, the DDP is performed by creating five sub-groups corresponding to each of the five datasets. The available GPUs are distributed evenly among the sub-groups.
The reported time represents the average over three training epochs.

As training is performed in batches of data, we refer to the local batch size as the size of a data batch on each GPU, whereas the effective batch size refers to the aggregated batch size across all GPUs. For example, if the local batch size on 8 GPUs in a test is 32, the effective batch size is 256.
For weak scaling tests, the local batch size per GPU is kept constant as we vary the number of GPUs to ensure that each GPU handles the same amount of work as we scale the training. On the other hand, for the strong scaling tests, we fix the effective batch size and vary the local batch size as we add more GPUs.

The weak scaling results show that the performance of MTL-par and MTL-base is similar on Frontier except for the lower batch size of 160. The increase in runtime across different GPU counts is due to the increase in communication overhead of gradient synchronization during the backward phase of training, which scales linearly with the increase in compute resources. We observe platform-specific scaling behavior, e.g., MTL-base shows lower runtime as compared with MTL-par on Perlmutter, indicating optimal performance for Perlmutter when a model fits completely in the GPU memory.

The strong scaling results show that MTL-par scales much better as compared with MTL-base at larger scale on both Frontier and Perlmutter. We observe near optimal scaling on both Frontier and Perlmutter up to 320 GPUs for larger values of the effective batch size. This shows the advantages of using parallel multi-task learning at large scale. In general, we observe higher variability in performance on Aurora as compared with the other two systems.

\section{Conclusion and Future Work}
\label{sec:conclusion}
We proposed a new approach for model parallelism customized on response to the need for computational readiness to enable computationally efficient, scalable, and stable pre-training of GFMs on a progressively increasing number of multi-source, multi-fidelity datasets released by different research groups. Our model parallelism is called multi-task parallelism and is specific to MTL architectures, where every branch of MTL is used to process data from a single data source. Relying on the algorithmic independence of the data processing from each MTL head, our multi-task parallelism approach assigns different branches of the MTL architecture to different GPUs, each executing concurrently the forward pass of the respective MTL heads during the training. 

We implemented this multi-task parallelism in HydraGNN, an open-source scalable GNN architecture suited for scalable data management training on supercomputing architectures, and pre-trained the GFM on 5 open-source atomistic scale datasets with over 24 million atomistic structures that include both organic and inorganic compounds an that cover over two-thirds of the natural elements of the periodic table. To ensure efficient scaling on large volumes of data, we also integrated our multi-task parallelism with DDP to form a 2D parallelization schema. The numerical results show that MTL is an effective training methodology to stabilize the GFM pre-training and the model maintains high predictive accuracy and transferability across broad regions of the chemical space. 
We tested the scaling performance of our multi-task parallelism integrated with DDP on three US-DOE supercomputers: the petascale system NERSC=Perlmutter, the exascale system ALCF-Aurora, and the exascale system OLCF-Frontier, using using up to 1,920 GPUs. Strong and weak scaling tests show that our approach enables effective scaling, due to the multi-task parallelism which allows to replace global communications that exchange large messages with local communications, each transmitting messages of smaller size. 
 
Future work will be dedicated to expanding the set of datasets reaching up to 359 million atomistic structures that cover all the natural elements of the periodic table, as we will illustrate the efficacy of our pre-trained on a broad class of downstream tasks.

\section*{Acknowledgments}
This research was sponsored by the Artificial Intelligence Initiative through the Laboratory Directed Research and Development (LDRD) Program at Oak Ridge National Laboratory (ORNL), managed by UT-Battelle, LLC, for the U.S. Department of Energy (DOE) under contract DE-AC05-00OR22725. It used resources of (i) the Oak Ridge Leadership Computing Facility, supported by the DOE Office of Science under the same contract (Director’s Discretionary award LRN070); (ii) the Argonne Leadership Computing Facility, a DOE Office of Science user facility at Argonne National Laboratory, supported by the Office of Science – Advanced Scientific Computing Research Program under contract DE-AC02-06CH11357 (Director’s Discretionary award HydraGNN); and (iii) the National Energy Research Scientific Computing Center (NERSC), a DOE Office of Science user facility, through award “GenAI@NERSC” ASCR-ERCAP0031171.

\bibliographystyle{splncs04}
\bibliography{references}

\end{document}